\documentclass[review]{elsarticle}
\usepackage{lineno,hyperref}
\usepackage{multirow}
\usepackage{subfigure}
\usepackage{chngpage}
\usepackage{array}
\usepackage{booktabs}
\usepackage[american]{babel}
\usepackage{microtype}
\usepackage{setspace}
\usepackage{xcolor}
\newcommand{\revision}[1]{\textcolor{black}{#1}}

\newcommand{\thirdvis}[1]{\textcolor{black}{#1}}

\modulolinenumbers[5]

\journal{Journal of Knowledge-Based Systems}

\begin{document}

\begin{frontmatter}

\title{SRQA: Synthetic Reader for Factoid Question Answering}

\author[address1,address2,address3]{Jiuniu Wang}
\ead{wangjiuniu16@mails.ucas.ac.cn}
\author[address1,address2]{Wenjia Xu}
\author[address1]{Xingyu Fu}
\author[address1]{Yang Wei}
\author[address1]{Li Jin}
\author[address1]{Ziyan Chen}
\author[address1]{Guangluan Xu}
\author[address1]{Yirong Wu}
%\cortext[mycorrespondingauthor]{Corresponding authors}

\address[address1]{Key Laboratory of Network Information System Technology (NIST), \\Institute of Electronics, Chinese Academy of Sciences, Beijing, China}
\address[address2]{University of Chinese Academy of Sciences, Beijing, China}
\address[address3]{City University of Hong Kong, Hong Kong}

%{\color{blue} }
\begin{abstract}
The question answering system can answer questions from various fields and forms with deep neural networks, but it still lacks effective ways when facing multiple evidences. We introduce a new model called \textbf{SRQA}, which means \textbf{S}ynthetic \textbf{R}eader for Factoid \textbf{Q}uestion \textbf{A}nswering. This model enhances the question answering system in the multi-document scenario from three aspects: model structure, optimization goal, and training method, corresponding to Multilayer Attention (MA), Cross Evidence (CE), and Adversarial Training (AT) respectively. \thirdvis{First, we propose a multilayer attention network to obtain a better representation of the evidences. The multilayer attention mechanism conducts interaction between the question and the passage within each layer, making the token representation of evidences in each layer takes the requirement of the question into account. Second, we design a cross evidence strategy to choose the answer span within more evidences. We improve the optimization goal, considering all the answers' locations in multiple evidences as training targets, which leads the model to reason among multiple evidences. Third, adversarial training is employed to high-level variables besides the word embedding in our model. A new normalization method is also proposed for adversarial perturbations so that we can jointly add perturbations to several target variables. As an effective regularization method, adversarial training enhances the model's ability to process noisy data.} Combining these three strategies, we enhance the contextual representation and locating ability of our model, which could synthetically extract the answer span from several evidences. We perform SRQA on the WebQA dataset, and experiments show that our model outperforms the state-of-the-art models (the best fuzzy score of our model is up to 78.56\%, with an improvement of about 2\%).
\end{abstract}
\begin{keyword}
	\texttt{Question Answering}\sep \texttt{Multilayer Attention}\sep\texttt{Cross Evidence}\sep\texttt{Adversarial Training} 
\end{keyword}

\end{frontmatter}

%\linenumbers
\section{Introduction}\label{Introduction}

The Question Answering (QA) system aims to generate answers to users' questions, so the system needs to search for answers within the related passages or knowledge bases. The research of the QA system can be divided into KBQA (Knowledge-Based Question Answering)~\cite{cui2017kbqa} and DBQA (Document-Based Question Answering)~\cite{rajpurkar2016squad,liu2019neural} according to the source of its answer. The factoid question answering in this paper belongs to DBQA  because it only uses documents to find the answer. \thirdvis{It mainly aims to} answer real-life fact-related questions that have certain answers, such as ``who is the first wife of Albert Einstein?''. These answers are often short text spans and mostly named entities. Factoid question answering, as a research platform for the QA system, is \thirdvis{appropriate} to design and evaluate DBQA models. \thirdvis{This task} provides a basis for model selection during constructing a real QA system. In this paper, the input of our model is several documents related to the question, which could either be retrieved from websites or provided by \thirdvis{Internet users}. And the final answer is the span in these documents located by the model.

In some applications such as search engines, a QA system has to face the problem of finding the answer within multiple candidate documents. The multiple documents, though providing more evidences, tend to make noise that might obstruct the seeking of the correct answer. Here \textit{evidence} means one or several sentences that contain the answer. It is challenging to get the appropriate representation and the answer span when the input is a long passage with multiple evidences. To solve this problem, we propose a novel model called Synthetic Reader for Factoid Question Answering (SRQA) with \thirdvis{the following} three strategies. The multilayer attention (MA) is applied to discover high-level information. The cross evidence (CE) strategy puts each answer span label of multiple evidences into optimization simultaneously. \thirdvis{So our model could} verify multiple evidences. Evidences splicing will increase the noise contained in the input, so we improve \thirdvis{the adversarial training (AT) method} to enhance the model's resistance to noise data. With the above strategies, SRQA can synthesize the semantic information contained in multiple evidences stably. Finally, we perform SRQA model on WebQA~\cite{li2016dataset}, a large scale real-world Chinese QA dataset.\footnote{The dataset and code can be downloaded from https://github.com/WangJiuniu/SRQA.}  

The representing ability is a key point for the QA system. Some previous works utilize pre-training to improve the representing ability, e.g., ELMo~\cite{Peters2018Deep} and BERT~\cite{devlin2018bert}. And the state-of-the-art models, such as DrQA~\cite{chen2017reading}, BIADF~\cite{seo2016bidirectional}, ReasoNet~\cite{shen2017reasonet}, R-Net~\cite{Group2017R} \thirdvis{and DynSAN~\cite{zhuang2019token}}, pay more attention on question-passage interactions. They have been proved to be effective in English machine reading comprehension datasets including CNN/DailyMail~\cite{hermann2015teaching} and SQuAD~\cite{rajpurkar2016squad}. They all uniformly use attention mechanism and pointer network~\cite{vinyals2015pointer} to predict the answer span. Following this idea, the attention mechanism is used in our model to deal with information redundancy caused by the long passage. We adopt the multilayer attention (MA) to focus on important words and their representation by refining information at each level. In the low-level (layers near the input of the model), the attention weight is highly affected by the similarity of the word embedding and lexical structure, e.g., {\itshape affix, part of speech}, which contains syntactic information. While in the high-level (layers near the output of the model), the attention variable could reflect the semantic information related to the passage and the question, since the contextual representation is fused during the interaction. For instance, \cite{huang2017fusionnet} shows the high-level representation of ``Alpine Rhine'' can be thought as ``Separating River'' in the \thirdvis{specific} passage.

\thirdvis{A simple idea is that the model's performance can be} improved by answering questions based on multiple evidences. Therefore we utilize the cross evidence (CE) strategy to seek the answer. The cross evidence (CE) strategy calculates the loss with all answer span labels from multiple evidences, which instructs our model to care more about the correlation between evidences. \thirdvis{ Besides,} the previous research found that the neural network lacks robustness~\cite{szegedy2013intriguing} though it achieves satisfying results on some tasks. It is easy to be disturbed by adversarial examples~\cite{jia2017adversarial}. In this condition, we adopt adversarial training (AT)~\cite{goodfellow2014explaining} as a regularization method to improve our model's generality and robustness. \thirdvis{Previous work mainly} applies adversarial perturbations on input signals~\cite{goodfellow2014explaining} or word embedding~\cite{miyato2016adversarial}, acting as a method to enhance the input data. While in this paper, \thirdvis{we blend} these perturbations into different model layers after adaptive normalization, covering the significant variables generated by question-passage interactions.

To sum up, our contributions can be summarized as follows:
\begin{itemize}
	\item To discover significant syntactic and semantic information from long passages, we apply the multilayer attention (MA) \thirdvis{mechanism} to each layer of our model. \thirdvis{Our model makes efficient interactions between the question and the passage, so it could emphasis on the tokens in evidences which are more relevant to the question.}
	\item The cross evidence (CE) strategy is utilized to read more relevant information between evidences, allowing these evidences from different documents to verify each other. \thirdvis{This strategy would guide the model to notice more about the consistency among evidences.}
	\item To adapt our model to the interference caused by multiple evidences, we improve the adversarial training (AT)  method \thirdvis{which is applied} on several variables. The adaptive normalized perturbations can be added to multiple variables at the same time. \thirdvis{As a regularization method, AT makes up for the weakness of the model, effectively improving the robustness of the model. So our model can be adapted to more complex application scenarios.}	The adversarial training not only enhances the information representation ability, but also promotes the answer locating ability of the whole model.
	\item We propose a novel neural network named SRQA, which gains the best \thirdvis{performance} on the WebQA dataset. The best fuzzy score of our model is 77.01\% for single evidence condition, and 78.56\% for multiple evidences condition.
\end{itemize}

The rest of this paper is organized as follows: Section 2 summarizes previous work related to our model, i.e., attention mechanism, multiple evidences, and adversarial training. In Section 3, we detailed the structure of our model and explained the usage of cross evidence and the improvement of the adversarial training. Section 4 shows the effectiveness of the contributions through the statistics and the case study of the experimental results. Section 5 concludes the whole paper and discusses the practical advantages and future research suggestions.

\section{Related Work}

The Question Answering (QA) System~\cite{ravichandran2002learning} is an advanced form of information retrieval system \thirdvis{aiming to answer} users' questions in an accurate and concise natural language. For example, Watson~\cite{high2012era} developed by IBM is an excellent QA system. The construction of the QA system \thirdvis{is often complex, and its} research could be mainly divided into two types \thirdvis{according answer searching strategies}. One is KBQA (Knowledge-Based Question Answering)~\cite{cui2017kbqa}, whose answer comes from the knowledge graph or knowledge base. This type of research mainly focuses on knowledge representation of nodes and relationships, as well as searching and matching in the knowledge graph. The other is DBQA (Document-Based Question Answering)~\cite{rajpurkar2016squad,liu2019neural}, whose answer comes from related documents. This type of research mainly focuses on the construction of multilayer neural networks \thirdvis{which could locate and fuse the answers. Some recent works~\cite{ding2019cognitive, bi2019incorporating, qiu2019machine} lead knowledge graphs or knowledge bases to assist the DBQA, which has made significant improvements in performance.}

Our study belongs to DBQA, and the related work is introduced in \thirdvis{the following} three aspects: attention mechanism, multiple evidences, and adversarial training.

\paragraph{Attention mechanism}
The attention mechanism has demonstrated success in a wide range of tasks. It was proposed by Bahdanau et al.~\cite{bahdanau2014neural} and was first applied to neural machine translation. Then it comes into use in many other tasks of the natural language processing. Similar to the main structure in Fig.~\ref{fig0}, a Neural QA system of DBQA \thirdvis{typically contains} three modules, i.e., embedding module, content encoding, and answer module. Attention near embedding module \thirdvis{(low-level, syntactic attention)} aims to attend the embedding from the question to the passage~\cite{chen2017reading}. Attention after context encoding \thirdvis{(high-level, semantic attention)} extracts the high-level representation in the question to augment the context. Self-match attention~\cite{tan2017s} is \thirdvis{often} adopted before answer module. It dynamically refines the representation by looking over the whole passage.

As shown in Table~\ref{tab1}, \thirdvis{the above} three different types of attention mechanisms are widely used in state-of-the-art models.  DrQA~\cite{chen2017reading} simply uses a bilinear term to compute the attention weights, so as to get word-level question-merged passage representation. FastQA~\cite{weissenborn2017making} combines features to calculate the word embedding attention weights. Match-LSTM~\cite{wang2016machine} applies LSTM~\cite{bahdanau2014neural} to extract feature from context and to concatenate attentions from two directions. A less memory attention mechanism is introduced in BIDAF~\cite{seo2016bidirectional} to generate bi-directional attention flow. R-Net~\cite{Group2017R} extends self-match attention to refine information over context. SAN~\cite{liu2017stochastic} adopts self-match attention and uses stochastic prediction dropout to predict the answer during the training process. Huang et al.~\cite{huang2017fusionnet} summarizes previous research and proposes fully-aware attention to fuse different representations over the whole model. \thirdvis{DynSAN~\cite{zhuang2019token} handles cross passage attention through align and uses dynamic self-attention to model long-range dependencies.} To better extract features from the passage, we utilize three methods to calculate attention weights. It helps our model to interchange information between questions and passages frequently, so as to notice important words among them.

\begin{table}
	\centering
	\caption{An outline of attention mechanism used in state-of-the-art architectures.}\label{tab1}
	\resizebox{\textwidth}{18mm}{
	\begin{tabular}{|c|c|c|c|}
		\hline
		Model &  Syntactic attention  &  Sematic attention  &  Self-match attention \\
		\hline
		DrQA~\cite{chen2017reading} & $\surd$ &  &  \\
		FastQA~\cite{weissenborn2017making}& $\surd$ &  &  \\
		Match-LSTM~\cite{wang2016machine}&  & $\surd$ &  \\
		BIDAF~\cite{seo2016bidirectional}&  & $\surd$ &  \\
		R-Net~\cite{Group2017R}&  & $\surd$ & $\surd$ \\
		SAN~\cite{liu2017stochastic}&  &  & $\surd$\\
		FusionNet~\cite{huang2017fusionnet}& $\surd$ & $\surd$ & $\surd$ \\
		\thirdvis{DynSAN~\cite{zhuang2019token}}& & $\surd$ & $\surd$ \\
		\hline
	\end{tabular}
	}
\end{table}

\paragraph{Multiple Evidences}

Multiple evidences could facilitate similarity comparison and answer selection in the QA system. Recently, some datasets with multiple evidences have also \thirdvis{concerned by} researchers, such as MS-MARCO~\cite{nguyen2016ms} and DuReader~\cite{he2017dureader}. \thirdvis{Some models, e.g., S-Net~\cite{tan2017s}, Full Re-ranker~\cite{wang2018evidence}, V-Net~\cite{Multi-Passage}, RE$^3$QA~\cite{hu2019retrieve}, and DynSAN~\cite{zhuang2019token}, focus on generating answers with multiple evidences.} These models aim to locate answers within multiple long paragraphs. 

It is believed that there are mainly two ways to handle multiple evidences, i.e., answer re-ranking and evidence concatenating. For answer re-ranking, one candidate answer is generated from each evidence, and the best answer is finally selected by sorting \thirdvis{the confidence score of} these candidates. \thirdvis{Several models, such as V-Net~\cite{Multi-Passage}, Full Re-ranker~\cite{wang2018evidence}  and RE$^3$QA~\cite{hu2019retrieve}, have mechanisms of scoring and sorting answers.} Evidence concatenating puts multiple evidences (or their vector representations) together near the embedding module, and \thirdvis{inputs the concatenating} evidence to the model after normalization. \thirdvis{Typical models with evidence concatenating are S-Net~\cite{tan2017s}, Neural Cascades model~\cite{swayamdipta2018multimention} and DynSAN~\cite{zhuang2019token}.}  S-Net contains a Passage Ranking which measures the relevancy between several evidences and the question. Neural Cascades model uses feed-forward networks to synthesize multiple answers in a long document. \thirdvis{DynSAN proposed a delicate dynamic self-Attention block to align the variable from each passage.} These concatenating models mainly operate on different sentences from the same document. They do not have special designs for sentences from multiple related documents. In this paper, we utilize cross evidence (CE) strategy based on multiple evidences spliced from \thirdvis{different} documents. During \thirdvis{each training step}, our model considers every golden answer of evidences, allowing the evidences to verify each other according to their semantic information.

\paragraph{Adversarial Training}
Szegedy et al.~\cite{szegedy2013intriguing} found that the deep neural network might make mistakes when adding small worst-case perturbations to its input. Many models cannot defend the perturbations, including \thirdvis{the} state-of-the-art neural networks such as CNN~\cite{krizhevsky2012imagenet} and RNN~\cite{cho2014learning}. In recent years, there are several methods for regularizing the parameters and features of deep neural networks during training. For example, by randomly dropping units, dropout~\cite{JMLR:v15:srivastava14a} is widely used as a simple way to prevent neural networks from overfitting. 

Adversarial training (AT)~\cite{goodfellow2014explaining} is a kind of regularizing learning algorithm. It was first proposed as a fine-tuning method for image classification. By adding perturbations to input signals during training, the neural network could \thirdvis{defend the attack} of adversarial example. Miyato et al.~\cite{miyato2016adversarial} first adopted AT on text classification. They add perturbations to word embedding and obtain similar benefits with that in image classification. AT was also utilized to Relation Extraction by Wu~\cite{wu2017adversarial}. In order to resist the noise interference leading by multiple evidences, we \thirdvis{employ} AT to several target variables within our model.

\section{Proposed Model}
\begin{figure}
	\centering
	\includegraphics[width=6cm, height=5.5cm]{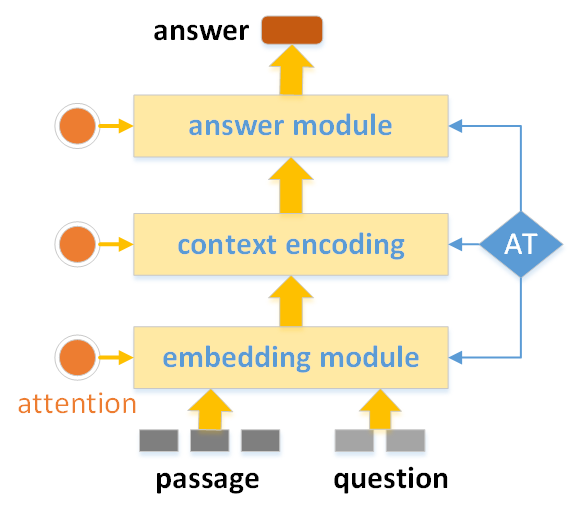}
	\caption{The overall structure of our model. The attention mechanism and adversarial training (AT) can be utilized in each module. The passage could represent either single evidence or a splice of evidences depending on the input data.} \label{fig0}
\end{figure}

According to Fig.~\ref{fig0}, the overall structure of our model consists of three main modules, including embedding module, context encoding, and answer module. Multilayer attention, as well as adversarial training, is applied in each module of our model. The goal of our model is to get the answer to a specific question $Q$ based on the related passage $P$. Our model first calculates the likelihood of being the start or end position of the answer for each token in the passage. The text span is then determined based on several constraints as the final output answer. The $Q$ and $P$ can be represented as:
 \begin{equation}
 Q = \{ {q_1},{q_2},..,{q_J}\},\textbf{ } P = \{ {p_1},{p_2},...,{p_T}\}
 \end{equation}
where $q_i$, $i \in \{1,2,\dots,J\}$ represents the $i$-th token in the question, and $p_j$, $j \in \{1,2,\dots,T\}$ represents the $j$-th token in the passage. In this paper, the uppercase letter indicates the whole variable, and the lowercase letter with a subscript indicates the subordinate variable corresponding to a single token. In our model, $P$ can be either single evidence or a splice of multiple evidences. These evidences come from the paragraphs retrieved from websites and the community QA product (i.e., Baidu Zhidao).

\subsection{Multilayer Attention}
\begin{figure}[!t]
	\centering
	\includegraphics[width=12cm , height=8.5cm]{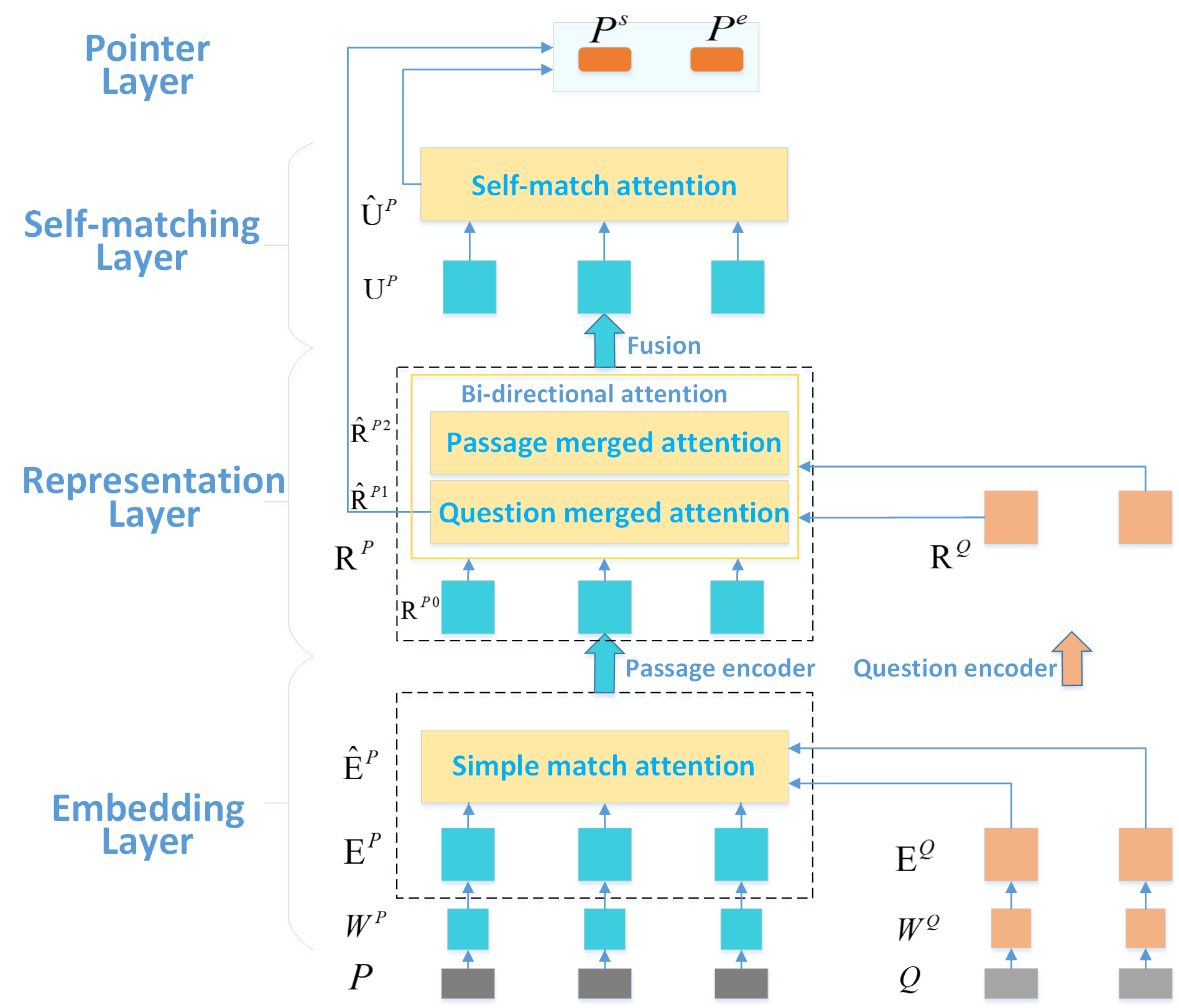}
	\caption{Multilayer attention structure of our model. The dotted box represents the concatenate operation. $P$ denotes Passage, $Q$ denotes Question, $W$ denotes character vector, $E$ denotes word vector, $R$ denotes the variable in Representation Layer, $U$ denotes the variable in Self-matching Layer, $P^s$ and $P^e$ denote answer points, variables with $\hat{ }$ is related to attention mechanism.} \label{fig1}
\end{figure}

As depicted in Fig.~\ref{fig1}, our multilayer attention structure can be decomposed into four layers: Embedding Layer, Representation Layer, Self-matching Layer, and Pointer Layer. Identical with Fig.~\ref{fig0}, Embedding Layer serves as embedding module, Representation Layer and Self-matching Layer belong to context encoding, and Pointer Layer works as answer module. \thirdvis{We apply} three attention methods to different layers. Simple match attention is adopted in Embedding Layer, extracting syntactic information between the question and the passage. In Representation Layer, bi-directional attention raises the representation ability by linking and fusing semantic information from the question and the passage. Finally, we adopt self-match attention to refine overall contextual representation in Self-matching Layer. 

\subsubsection{Embedding Layer}
\paragraph{Input Vectors} We use randomly initialized character embeddings to represent text. Firstly, each token in $P$ and $Q$ is represented as several character indexes. Afterward, each character is mapped to a high-density vector space (character vectors ${W^P}$ and ${W^Q}$). In order to get a fixed-size vector for each word, 1D max pooling is used to merge character vectors into word vector (${E^P}$ and ${E^Q}$). Simple match attention is then applied to match word-level information, which can be represented as follows: 
\begin{equation}
\hat E^P = SimAtt(E^P,E^Q)
\end{equation}
where $SimAtt(\cdot)$ denotes the function of simple match attention. 

\paragraph{Simple Match Attention} Here we additionally describe the simple match attention $SimAtt(\cdot)$ used in our model. Given two sets of vector ${V^A} = \{ v_1^A,v_2^A,\ldots,v_N^A\} $ and ${V^B} = \{ v_1^B,v_2^B,\ldots,v_M^B\} $. Now we synthesize information from ${V^B}$ for each vector in ${V^A}$. Firstly we get the attention weight $\alpha _{ij}$ of $i$-th token in $A$ and $j$-th token in $B$ by 
\begin{equation}
\alpha _{ij} =softmax(\exp (<v_i^A,v_j^B>))
\end{equation}
where $<>$ represents inner product. Then the sum for every vector in $V^B$ is weighted by $\alpha _{ij}$ to get the attention representation $\hat v_i^A = \sum\limits_j {\alpha _{ij}v_j^B}$. Attention variable $\hat V^A$ can be denoted as $\hat V^A = \{ \hat v_1^A,\hat v_2^A,\ldots,\hat v_N^A\}  = SimAtt(V^A,V^B)$.

\subsubsection{Representation Layer}
To better extract semantic information, we utilize RNN encoders to produce high-level representation $R^Q=\{r_1^Q,\ldots,r_J^Q\}$ and $R^{P0}=\{r_1^{P0},\ldots,r_T^{P0}\}$ for all the tokens in the question and the passage respectively. The encoders are made up of bi-directional Simple Recurrent Unit (SRU)~\cite{lei2017training}, which can be represented as follows:
\begin{equation}
r_t^P = BiSRU(r_{t - 1}^P,[e_t^P;\hat e_t^P]), \textbf{ }
r_j^Q = BiSRU(r_{j - 1}^Q,e_j^Q)
\end{equation}

\paragraph{Bi-directional Attention} Bi-directional attention is applied in this layer to combine the semantic information between the question and the passage. Similar to the attention flow layer in BIDAF, we compute \textit{question merged attention} ${\hat R^{P1}}$ and \textit{passage merged attention} ${\hat R^{P2}}$ as bi-directional attention. The similarity matrix is firstly computed by ${S_{ij}} = \beta (r_i^P,r_j^Q)$, we choose linear function
\begin{equation}
\beta (r_i^P,r_j^Q) = {W_{(S)}}^T[r_i^P;r_j^Q;r_i^P \cdot r_j^Q]
\end{equation}
where ${W_{(S)}}$ is the trainable parameters, $ \cdot $ is element-wise multiplication, $;$ means vector concatenation across row.

{\itshape Question merged attention ${\hat R^{P1}}$} signifies which question tokens are more relevant to each passage token. Question merged attention weight (the $i$-th token in the passage versus a certain token in the question) is computed by ${a_{i:}} = {\rm{softmax(}}{{\rm{S}}_{i:}}{\rm{)}} \in {{\rm{R}}^J}$. Each attended question merged vector can be denoted as ${\hat r_i}^{P1} = \sum\nolimits_j {{a_{ij}}r_j^Q} $. Thus we get the question merged attention $\hat R^{P1}=\{\hat r_1^{P1},\ldots,\hat r_T^{P1}\}$.

{\itshape Passage merged attention ${\hat R^{P2}}$} signifies which passage tokens have the closer similarity to each question token and hence critical for answering the question. The attended passage-merged vector is ${\tilde R^{P2}} = \sum\nolimits_i {{b_i}r_i^P} $, where $b = {softmax}(ma{x_{col}}(S))$ and $b \in {R^T}$, the maximum function $ma{x_{col}}()$ is performed across the column. Then ${\tilde R^{P2}}$ is tiled $T$ times to ${\hat R^{P2}} \in {R^{2d \times T}}$, where $d$ is the length of hidden vectors and $T$ is the number of passage tokens.

\subsubsection{Self-matching Layer}
The above bi-directional attention representation, i.e., ${\hat R}^{P1}$ and ${\hat R^{P2}}$, is concatenated with word representation $R^{P0}$ to generate the attention representation ${\hat R^P}$, denoted as follows:
\begin{equation}
{\hat R^P} = [{\hat R^{P1}};{\hat R^{P2}};{R^{P0}}]
\end{equation}
Then we use a bi-directional SRU as a Fusion to fuse information, which can be represented as $u_t^P = BiSRU(u_{t - 1}^{P - 1},{\hat r_t}^P)$.
In order to consider the whole passage, we apply self-match attention in Self-matching Layer. Note that the function is the same as simple match attention, except that its two inputs are the same variable ${U^P}$:
\begin{equation}
\hat U^P = SimAtt(U^P,U^P)
\end{equation}

\subsubsection{Pointer Layer}
Pointer network is a sequence-to-sequence model proposed by Vinyals et al.~\cite{vinyals2015pointer} In Pointer Layer, we adopt pointer network to calculate the possibility of the start and end position for every token in the passage. Instead of using a bilinear function, we take a linear function (which is proved to be simple and practical) to get the probability of start position ${P^s}$ and end position ${P^e}$ as:
\begin{equation}
{P^s} = {softmax}({W_{Ps}}[\hat u_i^P;\hat r_i^{P1}]),\textbf{ } {P^e} = {softmax}({W_{Pe}}[\hat u_i^P;\hat r_i^{P1}; P^s])
\end{equation}

\paragraph{Training} During training, we minimize the cross-entropy of the start and end labels of answer span as
\begin{equation}
L(\theta ) = -\frac{1}{N}\sum\limits_k^N {(\log (P_{i_k^s}^s) + \log (P_{i_k^e}^e))}
\label{loss_1} 
\end{equation}
where $i_k^s$, $i_k^e$ are the predicted answer span for the $k$-th instance, $N$ is the batch size.

\paragraph{Prediction} We predict the answer span to be $i_k^s$, $i_k^e$ with the maximum $P_{{i^s}}^s + P_{{i^e}}^e$ under the constraint $0 < {i^e} - {i^s} \le L_{max}$. \thirdvis{Here $L_{max}$ means the maximum answer token length constrainted our model. $L_{max}$ should be larger than most of the token length of answers in dataset. And we set $L_{max}=10$ in this work.} 

\subsection{Cross Evidence}
Retrieved Evidence in the dataset, denoted as $\{Ev_1,\ldots, Ev_K\}$, is used as multiple evidences to assist in locating answers in this paper. The suitable evidences are spliced together in our model. When people search for the answer to a question through search engines, most of them prefer to view the short evidence. In the same way, we try to select the evidences that are short and contain the answers for our model.

The multiple evidences are spliced as a passage and processed by cross evidence (CE) strategy. Because of the splicing of multiple evidences, multiple answer span labels would appear in one passage. We propose CE to make multiple answer span labels work together, leading our model notice the common rule contained in the evidences. Because each spliced passage contains a different number of evidences, the number of labels for each sample may be different. For each batch, we first count the maximum of the evidence numbers as
\begin{equation}
C_{max}=max\{C_1,\ldots,C_N\}
\end{equation} 
where $C_i$ represents the number of evidences of the $i$-th sample \thirdvis{in the batch}. In order to facilitate the calculation of the loss function, a random supplement strategy is employed in each training step. 
For the sample with evidences number less than $C_{max}$, we copy its existing answer span labels \thirdvis{until the number become $C_{\max}$}. After the supplement, each training sample would have $C_{max}$ answer span labels, though some of these labels are repeated. Now, the loss function Eqs.(\ref{loss_1}) can be extended as the following formula when training,
\begin{equation}
{\rm{L(}}\theta {\rm{) =  - }}\frac{1}{{N\cdot{C_{max}}}}\sum\limits_{k = 1}^N {\sum\limits_{t = 1}^{{C_{\max}}} {(log(P_{i_{k,t}^s}^s) + log(P_{i_{k,t}^e}^e))} } 
\end{equation}
where $N$ is the batch size, ${ i_{k,t} ^s }$/${ i_{k,t} ^e }$ is the $t$-th start/end position label of the $k$-th sample, and ${P_{i_{k,t}^s}^s}$/${P_{i_{k,t}^e}^e}$ represents the predicted probability to be start/end position for ${ i_{k,t}^s}$/${ i_{k,t} ^e}$. In this way, instead of picking one answer span label in one step, we consider the start and end positions of every correct answer span simultaneously during training. Note that when testing, we only need to search for the best answer in the entire passage.

\subsection{Adversarial Training}
Adversarial training applies the worst-case perturbations on target variables. As it is shown in Fig.~\ref{fig2}, $X$ denotes the target variable, and $\theta $ denotes the parameters of the model. Different from the previous work~\cite{miyato2016adversarial,wu2017adversarial}, $X$ can be set as each variable in our model without interfering with each other because of our perturbation normalization. Adversarial training adds adversarial loss function ${L_{adv}}(X;\theta )$ to the original loss $L(\theta )$. 
The equation of ${L_{adv}}(X;\theta )$ is described as follows:
\begin{equation}
{L_{adv}}(X;\theta ) = L(X + {r_{adv}};\theta ) ,\textbf{ } {r_{adv}} = \arg \mathop {\max }\limits_{||r|| \le \varepsilon ||X|| } L(X + r;\hat \theta )
\label{e9} 
\end{equation}
where $r$ is a perturbation on the target variable and $\hat \theta $ is a fixed copy to the current parameters. When optimizing parameters, the gradients should not propagate through ${r_{adv}}$. One problem is that we cannot get the exact value of ${r_{adv}}$ simply following Eqs.(\ref{e9}), since the computation is intractable. Following Goodfellow et al.~\cite{goodfellow6572explaining}, we approximate the value of ${r_{adv}}$ by linearizing $L(X;\hat \theta )$ near $X$ as:

\begin{figure} %[tb]
	\centering
	\includegraphics[width=6cm, height=5.2cm]{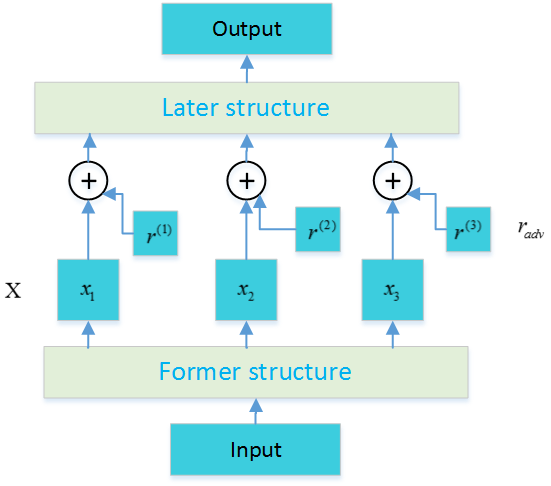}
	\caption{The computation flow chart of Adversarial Training. $X$ denotes target variable, ${r_{adv}}$ denotes adversarial perturbation. The input of the model is mapped into the target variable $X$ by the former structure. And then later structure generates the output based on the target variable $X$ combined with adversarial perturbation ${r_{adv}}$.} \label{fig2}
\end{figure}

\begin{equation}
{r_{adv}} = \varepsilon X \otimes \frac{g}{{||g||}} ,\textbf{ } g = {\nabla _X}L(X|\hat \theta )
\label{e10} 
\end{equation}
where $|| \cdot ||$ denotes the norm of variable $ \cdot $, $\otimes$ means elementwise product, and $\varepsilon $ is an intensity constant to adjust the relative norm of $r_{adv}$. In order to make the adversarial training method work on multiple variables at the same time, we use $X$ as a factor of the product, making $r_{adv}$ more similar to $X$. This reduces the perturbation's effects on other variables during training. ${r_{adv}}$ is different for each training sample and training step. Its direction is decided by the elementwise product of $X$ and $g$. When applying cross evidence strategy, we calculate $r_{adv}$ for each evidence respectively to reduce mutual interference between evidence.

\section{Experiments}
In this section, we evaluate our model on the WebQA dataset. Outperforming the baseline model in the original paper~\cite{li2016dataset} and several state-of-the-art model~\cite{chen2017reading, seo2016bidirectional, Group2017R}, we obtain higher fuzzy score with multilayer attention (MA), cross evidence (CE) and adversarial training (AT). For Annotated Evidence, we improve the fuzzy score from 73.50\% to 77.01\%. For Retrieved Evidence, the fuzzy score raises from 74.69\% to 78.56\%. 

\subsection{Dataset and evaluation metrics}

\begin{table}
	\centering
	\caption{An example of the WebQA dataset. The correct answer is in bold font.}\label{tab0}
	\begin{tabular}{|c|p{8cm}|}
		\hline
		Question & Who is the first wife of Albert Einstein?  \\
		\hline
		Annotated Evidence & Einstein married his first wife \textbf{Mileva Marić} in 1903.  \\
		\hline
		\multirow{5}*{Retrieved Evidence} &  Albert's three children were from his relationship with his first wife,
		Mileva Marić. His daughter Lieserl was born a year before they married. \\
		\cline{2-2}
		& \textbf{Mileva Marić}, Albert Einstein's first wife, was born 141 years ago on Dec. 19, 1875.(2016)\\
		\hline
		Answer & \textbf{Mileva Marić} \\
		\hline
	\end{tabular}
\end{table}

\begin{table}
	\centering
	\caption{Statistics of WebQA dataset.}\label{tab3}
	\begin{tabular}{|c|r|r|r|r|r|r|}
		\hline
		\multirow{2}*{Dataset} & \multicolumn{2}{c|}{Question}  & \multicolumn{2}{c|}{Annotated Evidence} & \multicolumn{2}{c|}{Retrieved Evidence}\\
		\cline{2-7}
		& \multicolumn{1}{c|}{\#} & \multicolumn{1}{c|}{word\#} & \multicolumn{1}{c|}{\#} & \multicolumn{1}{c|}{word\#} & \multicolumn{1}{c|}{\#} & \multicolumn{1}{c|}{word\#}\\
		\hline
		Train & 36,145 & 374,500 & 140,897 & 10,757,652 & 171,838 & 7,233,543  \\
		Validation & 3,018 & 36,666 & 5,412 & 233,911 & 60,351 & 3,633,540 \\
		Test & 3,024 & 36,815 & 5,445 & 234,258 & 60,465 & 3,620,391 \\
		\hline
	\end{tabular}
\end{table}

Table~\ref{tab0} shows an example of the WebQA dataset.
This dataset is made up of Question, Annotated Evidence, Retrieved Evidence, and Answer. Different from SQuAD~\cite{rajpurkar2016squad}, questions in WebQA are from user queries in search engines, and its passages are from web pages. WebQA provides a number of short evidence for each question in Retrieved Evidence. So \thirdvis{we use this dataset to} test the answer locating ability of our model under multiple evidences. 

Now we explain the statistical characteristics of this dataset. The statistic information is shown in Table~\ref{tab3}. We train and evaluate our model on Annotated Evidence and Retrieved Evidence respectively. For each question, there is an annotated evidence with a golden answer. Meanwhile, the question is accompanied by several pieces of Retrieved Evidence. In this paper, we prefer to use the short evidences which contain the golden answer. 

The evaluation of the QA systems is essential~\cite{RODRIGO201783}. Most golden answers in WebQA are simple nouns, such as person name, location, and time, so that 95.60\% of the golden answers are less than five tokens. Thus we decide to measure the accuracy of predicted answer directly, instead of using approximate evaluations such as BLEU and ROUGE. By comparing predicted answers with golden answers, the model performance can be evaluated by precision (\textbf{P}), recall (\textbf{R}) and F1-measure (\textbf{F1}): 

\begin{equation}
\textbf{P} = \frac{{|C|}}{{|A|}},\textbf{ R} = \frac{{|C|}}{{|Q|}},\textbf{ F1} = \frac{{2PR}}{{P + R}}
\end{equation}

\noindent where $|C|$ is the number of correctly answered questions, $|A|$ is the number of predicted answers given by the model, and $|Q|$ is the number of all questions. 
The same answer in WebQA may have different surface forms, such as “Beijing” v.s. “Beijing city”. In order to measure the correct answer reasonably, we use two ways to count correctly answered questions, which are referred to as {\itshape Strict} and {\itshape Fuzzy}. Strict matching means the predicted answer is \thirdvis{exactly} the same with the golden answer; Fuzzy matching means the predicted answer could be a synonym of the golden answer.

\subsection{Model details}
In our model, we use randomly initialized 64-dimensional character embedding, and then the character vectors in the same word are fused into word embedding with the same dimension by 1D max pooling. The hidden vector length $d$ is set to 100 for all layers. We utilize the 4-layer passage encoder and question encoder. The Fusion SRU is set to 2-layer. We also apply dropout between layers, with a dropout rate of $0.1$. The model is optimized using AdaDelta with a batch size of 64 and an initial learning rate of $0.01$. During training, we set the maximum answer length as 10. The maximum passage length for single evidence condition is 80, while that for multiple evidences condition is 140. Hyper-parameter $\varepsilon $ is selected from $10^{-5}$ to $10^{-1}$ according to the model performance on WebQA validation dataset. Our SRQA(MA+CE+AT) model takes about 8 hours to train on a single NVIDIA P100 GPU.

\subsection{Main results}

The main results are shown in Table~\ref{tab3_1}. \thirdvis{Several state-of-the-art models are evaluated on the WebQA dataset, including simple baselines (LSTM+softmax and LSTM+CRF)~\cite{li2016dataset}, DrQA~\cite{chen2017reading}, BIDAF~\cite{seo2016bidirectional}, R-net~\cite{Group2017R} and BERT~\cite{devlin2018bert}. 
Simple baselines are based on LSTMs and use sequence labeling to mark answers. The simple use of LSTM leads their limited text representation ability, and their sequence label method causes the low precision, so that their fuzzy score under single evidence is lower than 70\%. The remaining models are implemented based on the attention mechanism and the pointer network. DrQA, BIDAF, and R-net each proposed innovative attention methods. In particular, BIADF's bi-directional attention flow has become a foundational work in this field. However, due to they use fewer types and levels of attention mechanisms than us, their strict score and fuzzy score are lower than our SRQA(MA). BERT applies transformer~\cite{vaswani2017attention} as its attention mechanism, and is trained based on ultra-large-scale corpora, so it achieves competitive performance with fuzzy score 75.58\% for single evidence and 76.83\% for multiple evidences.} 

Corresponding to the contributions of this paper, we evaluate the performance of models under different conditions. We apply our cross evidence (CE) strategy under Retrieved Evidence since there is only one annotated evidence but several retrieved evidences for one question. SRQA(MA) denotes the multilayer attention (MA) model which does not apply adversarial training (AT). SRQA(MA+RN) denotes the contrast experiment which replaces adversarial perturbations with Random Gaussian Noise (RN) with a scaled norm. SRQA(MA+AT) is the model with adversarial training. Baseline models utilize the sequence label method to mark the answer, while others adopt the pointer network to locate the answer. Sequence label methods, such as LSTM+softmax and LSTM+CRF, can mark several answers for one question, leading to high recall (R) but low precision (P). So we adopt the pointer network to generate one answer for each question. In this condition, evaluation metrics (P, R, F1) are equal. Thus we can use this {\itshape score} to evaluate our model. Besides, fuzzy evaluation is closer to the requirement of \thirdvis{daily use}, so we mainly focus on {\itshape fuzzy score}.

\begin{table}
	%\begin{adjustwidth}{-.5in}{-.5in}  
	\caption{Evaluation results on the test dataset of WebQA. In this table, MA denotes Multilayer Attention; AT denotes Adversarial Training; RN denotes Random Noise; CE denotes Cross Evidence. The scores under Retrieved Evidence come from the model with Cross Evidence (CE). \thirdvis{Note that} the Precision and Recall for sequence label methods are different, while those for answer point methods are the same.}\label{tab3_1}
	\centering
	 \resizebox{\textwidth}{22mm}{
	\begin{tabular}{|c|c|c|c|c|c|c|c|c|c|c|}
		\hline
		\multicolumn{2}{|c|}{\multirow{3}*{Model}} & \multicolumn{6}{c|}{Annotated Evidence}  & \multicolumn{3}{c|}{Retrieved Evidence (CE)} \\
		\cline{3-11}
		\multicolumn{2}{|c|}{}& \multicolumn{3}{c|}{Strict Score}  & \multicolumn{3}{c|}{Fuzzy Score} & \multicolumn{3}{c|}{Fuzzy Score}\\
		\cline{3-11}
		\multicolumn{2}{|c|}{}& P & R & F1 & P & R & F1 & P & R & F1\\
		\hline
		\multicolumn{2}{|c|}{LSTM+softmax} & 59.38 & 68.77 & 63.73 & 63.58 & 73.63 & 68.24 &69.75&	74.72&	72.15\\
		\multicolumn{2}{|c|}{LSTM+CRF} & 63.72 & \textbf{76.09} & 69.36 & 67.53 & \textbf{80.63} & 73.50&72.66&	76.83&	74.69 \\
		\multicolumn{2}{|c|}{\revision{DrQA~\cite{chen2017reading}}} & 69.62 & 69.62 & 69.62 & 72.86 & 72.86 & 72.86& 75.24&	75.24&	75.24\\
		\multicolumn{2}{|c|}{BIDAF~\cite{seo2016bidirectional}} & 70.04 & 70.04 & 70.04 & 74.43 & 74.43 & 74.43& 75.62&	75.62&	75.62\\
		\multicolumn{2}{|c|}{\revision{R-net~\cite{Group2017R}}} & 70.48 & 70.48 & 70.48 & 74.82 & 74.82 & 74.82& 76.06&	76.06&	76.06\\
		\multicolumn{2}{|c|}{\thirdvis{BERT~\cite{devlin2018bert}}} & 71.36 & 71.36 & 71.36 & 75.58 & 75.58 & 75.58& 76.83&	76.83&	76.83\\
		\hline
		\multirow{3}*{SRQA} & MA & 71.03 & 71.03 & 71.03 & 75.46 & 75.46 & 75.46 &77.23&	77.23&	77.23\\
		& MA+RN & 71.28 & 71.28 & 71.28 & 75.89 & 75.89 & 75.89& 77.84&	77.84&	77.84\\
		&MA+AT & \textbf{72.51} & 72.51& \textbf{72.51}&  \textbf{77.01} &77.01&  \textbf{77.01} &\textbf{78.56}&\textbf{78.56}&	\textbf{78.56}\\
		\hline
	\end{tabular}
	} 
	%\end{adjustwidth}
\end{table}

The models using the attention mechanism (DrQA, BIDAF, and R-net) tend to have a higher F1 score compared to the baseline model (LSTM+softmax, LSTM+CRF). R-net can reach 74.82\% for single evidence condition and 76.06\% for multiple evidences condition. Benefit from multilayer attention, SRQA(MA) gains 0.64\% and 1.17\% promotion in the fuzzy score under the above two conditions compared to R-net. It indicates that multilayer attention is useful. The model can find the correct answer more easily under the influence of cross evidence, so the fuzzy F1 of SRQA(MA+CE) can achieve 77.23\%. Because of adversarial training (AT), SRQA(MA+AT) get a promotion of 1.12\% in Fuzzy F1 comparing to SRQA(MA), and SRQA(MA+CE+AT) get a promotion of 1.33\% comparing to SRQA(MA+CE). It indicates that AT has strong adaptability since it could work under a variety of conditions.

To demonstrate the superiority of adversarial training over the addition of noise, we include \thirdvis{contrast} experiments which replaced adversarial perturbations with random perturbations from a Gaussian distribution. From the performance of SRQA(MA+RN) and SRQA(MA+AT) shown in Table~\ref{tab3_1}, we notice that adversarial training is superior to random noise. Theoretically, noise is a far weaker regularization than adversarial perturbations. An average noise vector is approximately orthogonal to the cost gradient in high dimensional input spaces, while adversarial perturbations are explicitly chosen to consistently increase the loss value. Random noise is used to replace the worst-case perturbations on each target variable, which only leads to slight improvement. This indicates it is AT that improves the robustness and generalization of our model.

\subsection{Contributions analysis}
In order to observe the achievement of our contributions, we analyze the test results of three corresponding models. They are SRQA(MA), SRQA(MA+CE) and SRQA(MA+CE+AT). As shown in Fig.~\ref{fig3_1}, among the overall $3,023$ test samples, 62.75\% of them are answered correctly by every model, and 11.09\% of them are answered incorrectly. \thirdvis{It can be seen from this figure that each method has a specific proportion of samples that are correctly answered by themselves, 3.94\% for SRQA(MA), 2.34\% for SRQA(MA+CE) and 3.04\% for SRQA(MA+CE+AT). While a larger proportion of samples are coincident. When considering the impact of CE strategy, we view SRQA(MA) and SRQA(MA+CE). The overlap between the two methods is 66.82\%, and their different percentages are 8.64\% and 10.41\%. When analyzing the AT strategy, we choose SRQA(MA+CE) and SRQA(MA+CE+AT). The results of these two models overlap with 70.82\%, and their different percentages are 6.61\% and 7.74\%. It shows that the performance of the model using these two strategies is generally stable. And CE makes the model results change more than AT does.}

In general, the CE and AT strategies have a positive influence on the QA system in view of the percentage of correct answers. It is because the CE strategy allows the model to search the answer from multiple evidences into account simultaneously, and AT can effectively reduce the interference caused by the redundant passage spliced from multiple documents.

\subsection{\thirdvis{Cases analysis}}
\begin{figure} %[tb]
	\centering
	\includegraphics[width=5.8cm, height=5.2cm]{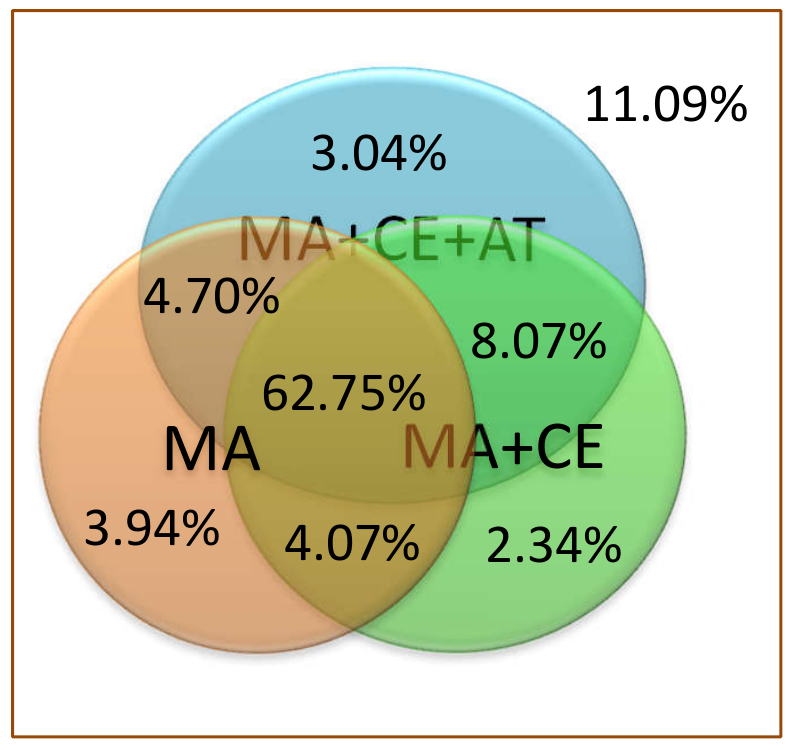}
	\caption{The Venn diagram of \thirdvis{fuzzy match result} of three main models (MA, MA+CE, MA+CE+AT) among $3,023$ test samples. The numbers in the figure indicate the \thirdvis{percentage of fuzzy match} in the WebQA test dataset. The actual proportion of each part is marked by the numerical label. For example, 11.09\% represents there are 11.09\% test samples not correctly answered by every model, while 62.75\% represents there are 62.75\% test samples correctly answered by all three models. In this figure, MA denotes Multilayer Attention; AT denotes Adversarial Training; CE denotes Cross Evidence.} \label{fig3_1}
\end{figure}

\thirdvis{In Table~\ref{tab3_2}, we list two cases to explain in detail. According to Fig.~\ref{fig3_1}, there are 19.06\% samples with a different result between SRQA(MA) and SRQA(MA+CE). Here we select one sample answered only wrongly by SRQA(MA) and another only correctly by SRQA(MA). For Sample 1, the answer from SRQA(MA) is \textit{carrier rocket}. It is wrong although the tokens are relevant to the problem, while the models with CE strategy can answer correctly. It is because SRQA(MA+CE) obtain more information from the Retrieved Evidence, and most of them mentioned \textit{Red East 1}. For Sample 2, \textit{Qufu} appears too many times in Retrieved Evidence. But it is only a city, not a province. So SRQA(MA) can get the correct answer \textit{Shandong Province}, but the models with CE strategy does not. So choosing the right model is critical when dealing with practical tasks. Only when the model is compatible with data and questions can it exert its maximum value.}

\thirdvis{The main purpose of the case study is to find the defects of the model and propose targeted improvements. For example, the maximum length of answers can be adjusted by observing whether the length of the extracted answer is too long or too short; the dimensions of the corresponding features can be adjusted by the sensitivity of the model to the case of letters, parts of speech, or meaning of words. Further, the hyperparameters of the model can also be adjusted by whether the attention values within the model are in line with expectations. In addition, if we care more about one or several wrongly answered questions in real business systems, a straightforward way to solve the problem is introducing or emphasizing similar training data. Take the wrong prediction in Sample 2 as an example, we could finetune the model with more questions about ``which province'', then it will be inclined to choose a province as the answer.}

\begin{table}
	\caption{\thirdvis{The result of two test samples.} The correct answers are in bold font.} 
	\label{tab3_2}
	\centering
	\resizebox{\textwidth}{30mm}{
	\begin{tabular}{|c|c|p{6cm}|p{6cm}|}
		\hline
		\multicolumn{2}{|c|}{} & Sample 1 & Sample 2 \\
		\hline
		\multicolumn{2}{|c|}{\multirow{2}*{Question}} & What is the name of the first artificial satellite launched in China?  & Which today's province Confucius was born in? \\
		\hline
		\multicolumn{2}{|c|}{\multirow{4}*{Annotated Evidence}}& On April 24th, 1970, China's long-distance carrier rocket successfully launched the first artificial satellite, \textbf{Red East 1}... &  Confucius was born in Changping Township of Lu State (today's Luyuan Village, Nanxin Town, southeast of Qufu City, \textbf{Shandong Province}).   \\
		\hline
		\multicolumn{2}{|c|}{\multirow{6}*{Retreived Evidence}} & On April 24, 1970, the first artificial earth satellite \textbf{Red East 1} ... was successfully launched by the 'Long March 1' carrier rocket. &  Qufu is the hometown of Confucius, located in the southwestern part of \textbf{Shandong Province}. \\
		\cline{3-4}
		\multicolumn{2}{|c|}{} & The first artificial satellite developed by our country is called \textbf{Red East 1}...&The hometown of Confucius is \textbf{Shandong} Qufu.   \\
		\hline
		\multirow{3}*{Answer} & MA& carrier rocket & \textbf{Shandong Province}\\
		& MA+CE& \textbf{Red East 1} & Qufu\\
		& MA+CE+AT& \textbf{Red East 1} & Qufu\\
		\hline
	\end{tabular}
	
	}
	
\end{table}

\begin{table}
	\caption{Comparison of different configurations of the basic model.} 
	\label{tab4}
	\centering
	\resizebox{\textwidth}{15mm}{
		\begin{tabular}{|c|c|c|c|c|}
			\hline
			\multirow{3}*{Configuration} & \multicolumn{2}{c|}{without CE}  & \multicolumn{2}{c|}{with CE}  \\
			\cline{2-5}
			&  Strict Score & Fuzzy Score& Strict Score & Fuzzy Score\\
			\hline
			SRQA(MA) \revision{basic} model & {\bfseries 71.03} & {\bfseries 75.46}  & {\bfseries 72.52} & {\bfseries 76.95}\\
			without Embedding Attention & 70.57 & 74.93 & 71.66 & 76.02\\
			without question merged Attention & 70.77&75.18 & 71.86 &76.27 \\
			without Bi-directional Attention & 70.63& 74.56& 71.72& 75.65 \\
			without Self-match Attention & 70.70& 75.23 & 71.79& 76.32\\
			\hline
		\end{tabular}
}
\end{table}

\subsection{Ablation on basic model structure}
Next, we investigate the ablation study on the structure of our basic model. From Table~\ref{tab4}, we can know that both the strict score and fuzzy score would drop when we omit any designed attention. It indicates that each attention layer in SRQA is essential. According to Table~\ref{tab4}, bi-directional attention is the most important in our model. Because if there is no bi-directional attention, the performance of our model will be reduced by more than 1\%.

\subsection{Adversarial Training on different target variables}

\begin{table}
	\caption{Comparison of Adversarial Training results on different target variables. The symbols in this table are corresponding with Fig.~\ref{fig1}.}\label{tab5}
	\centering
	\begin{tabular}{|c|c|c|c|c|}
		\hline
		\multirow{2}*{Target variable} & \multicolumn{2}{c|}{SRQA(MA)} & \multicolumn{2}{c|}{SRQA(MA+CE)} \\
		\cline{2-5}
		& Strict Score & Fuzzy Score &  Strict Score & Fuzzy Score\\
		\hline
		none (\revision{basic} model) & 71.03 & 75.46 & 72.58 & 77.03 \\
		${W^P}$& 71.95 & 76.62  & 72.82 & 77.44 \\
		${E^P}$ & 72.06 & 76.39 & 73.50 & 78.17 \\
		${\hat E^P}$& 71.32 & 75.92 &  73.61 &   77.94 \\
		${\hat R^{P1}}$ & 71.85 & 76.28 & 72.87 & 77.47\\
		${\hat R^{P}}$ & 71.56 & 76.42 & 73.40 & 77.83\\
		${W^P}$ and ${\hat R^P}$ & {\bfseries 72.51} &  {\bfseries 77.01} & 73.14 &   78.11\\
		${E^P}$ and ${\hat R^P}$ & 71.92 &  76.55 & {\bfseries 73.83} &  {\bfseries 78.56}\\
		\hline
	\end{tabular}
\end{table}

In this section, we evaluate the predicted result when applying adversarial training (AT) on different target variables. As it is shown in Table~\ref{tab5}, fuzzy score, as well as strict score, can be improved in different degrees by applying AT on each target variable. It indicates that AT can work as a regularizing method not only for word embeddings, but also for many other variables. Note that the score is significantly improved when applying AT on character embedding variable ${W^P}$, word embedding variable ${E^P}$ and attention variable ${\hat R^P}$. It reveals that AT can improve the representing ability for both inputs and non-input variables. Furthermore, the model can achieve better performance when applying AT on several variables at the same time. SRQA(MA) obtains the best result when applying AT on both ${W^P}$ and ${\hat R^P}$, while the best result of SRQA(MA+CE) is obtained when applying AT on ${E^P}$ and ${\hat R^P}$.

\begin{figure}
	\centering
	\includegraphics[width=9.5cm, height=5.5cm]{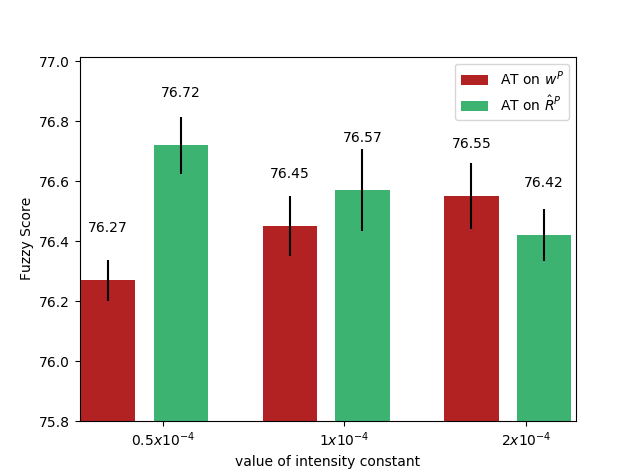}
	\vspace{-0.5em}
	\caption{Effect of intensity constant $\varepsilon $ when applying AT on passage word vector ${W^P}$ and passage Representation variable ${\hat R^P}$. This experiment is performed without CE.} \label{fig3}
\end{figure}

In order to measure the impact of intensity constant $\varepsilon $ in our model, AT is evaluated on two target variables (${W^P}$ and ${\hat R^P}$) under different $\varepsilon $ value. As shown in Fig.~\ref{fig3}, we repeat the experiment 3 times for each target variable on each constant $\varepsilon $, and get the average fuzzy score and its $std.$ error. For AT on attention variable ${\hat R^P}$, we obtain the best performance when $\varepsilon $ is $0.5 \times {10^{ - 4}}$; While for AT on character embedding variable ${W^P}$, we obtain the best performance when $\varepsilon $ is $2 \times {10^{ - 4}}$. It indicates we need larger adversarial perturbation for the low-level variable. This phenomenon could be explained in the following two different views. Firstly, ${W^P}$ and ${\hat R^P}$ are in different concept levels. ${W^P}$ contains syntactic meaning and represents as character embedding vectors. Most of the vectors can still hold original meaning under small perturbation because most points in embedding space have no real meanings. But ${\hat R^P}$ contains semantic meaning. Any perturbation on it would change its meaning. Thus our model is sensitive to the perturbation on ${\hat R^P}$. Secondly, ${W^P}$ and ${\hat R^P}$ are in different layers of our model. ${\hat R^P}$ is closer to the Pointer Layer, which could affect the output of the model and computation of loss function more directly. 
\begin{figure}[!h]
	\centering
	\subfigure[Fuzzy Score (test) under different training step.]{\label{fig4:a}
		\includegraphics[width=0.8\linewidth]{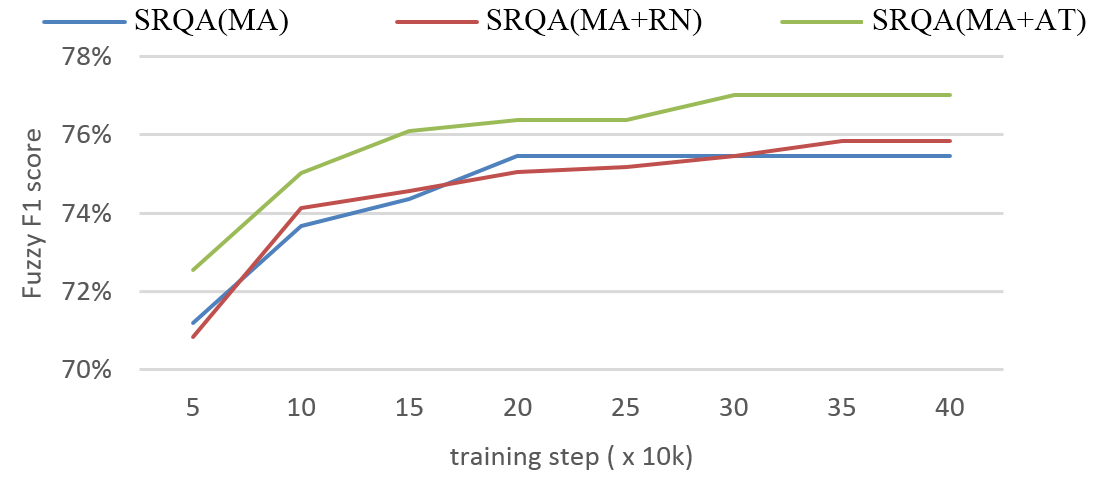}}
	
	\subfigure[Loss value (train) under different training step.]{\label{fig4:b}
		\includegraphics[width=0.8\linewidth]{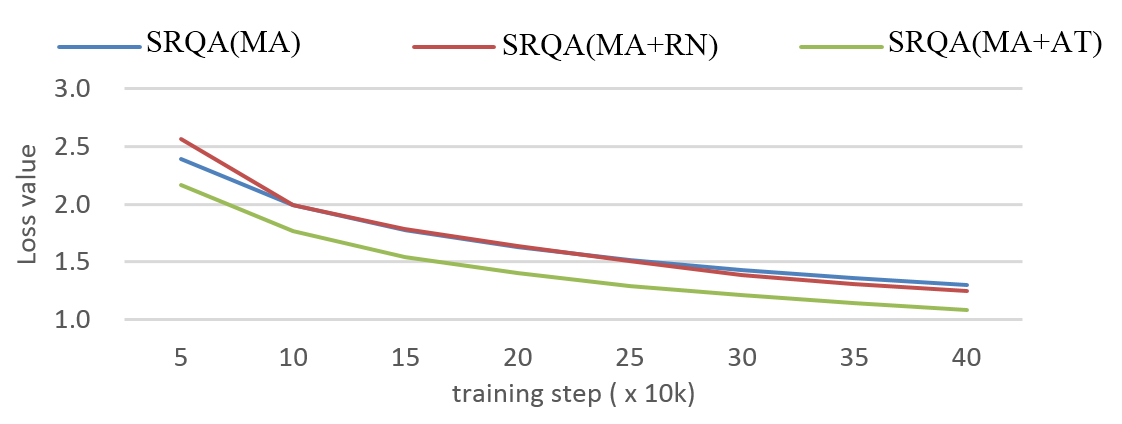}}
	\caption{Fuzzy Score and Loss value under different training step.}
	\label{fig4}
\end{figure}
\subsection{Effectiveness of Adversarial Training}

Afterward, we compare the performance of models with and without AT based on SRQA(MA). Fig.~\ref{fig4:a} shows the fuzzy score on the test dataset, and Fig.~\ref{fig4:b} shows the loss value on the training dataset under different configurations. The curves of SRQA(MA) and SRQA(MA+RN) are close to each other in both two subfigures. It indicates that random noise has a limited effect on our model. Within each training step, the fuzzy score of SRQA(MA+AT) is the highest, and its loss value is the lowest in Fig.~\ref{fig4}. It demonstrates that adversarial training can lead to better performance with less training steps.

\section{Conclusions}

The factoid question answering model could be a basic module in a question answering system. It aims to extract the text span from the passage to answer the question. In order to enhance the ability to handle multiple evidences, SRQA focuses on improving the previous work in three aspects, i.e., the model structure, the optimization goal, and the training methods. \thirdvis{With multilayer attention mechanism}, cross evidence strategy, and adversarial training method, our SRQA outperforms other state-of-the-art models on the WebQA dataset.
	
Multilayer attention aims to focus on the important words from the passage related to the question. This is beneficial to reduce the redundant content in the passage. Cross evidence strategy is also explored to take advantage of multiple evidences. Multiple answer span labels from each evidence are used together for training, which helps to take mutual verification from each other. Meanwhile, as a regularization method, adversarial training can be applied to almost every variable in the model under our well-designed normalization.

SRQA can find the answers to users' queries more accurately from multiple evidences, which could play an important role in the QA system and the search engine. In addition, the main contributions of this paper are actually about improving multi-source long text representation and model training. So they can be extended to many other natural language processing systems, such as sentiment analysis, text classification, natural language inference, named entity recognition, etc.

This paper focuses on the study of factoid question answering in the case of multiple evidences. So we paid limited effort to the word embedding and the language model, such as word2vec~\cite{mikolov2013distributed}, EMLo~\cite{Peters2018Deep}, BERT~\cite{devlin2018bert}, etc. Those methods \thirdvis{could} further enhance the text representation capabilities of the model. Another limitation is that we preferred to chose the short evidences and discarded the long one in our experiment.  While in practice, the long evidences could also benefit in finding the answer. It would be better to come up with a more reasonable evidence selection strategy.

In addition to improving the model in terms of text representation and evidence selection, future research could also lie in the following two aspects. The factoid question answering task is quite basic and simple. It is meaningful to design a model that could answer more complex questions with longer answers (e.g. ``how to make dumplings?''). Moreover, the answer selection and fusion method, related to generating the final answer based on the candidate answers from multiple evidences, is worth studying.

%\clearpage
\section*{References}

\bibliography{mybibtex}

\end{document}